\documentclass[letterpaper, 10 pt, conference]{ieeeconf} 

\IEEEoverridecommandlockouts                      

\usepackage{graphicx}
\usepackage{svg}

\usepackage{wrapfig}
\usepackage{booktabs, tabularx}
\newcolumntype{C}{>{\centering\arraybackslash}X}
\usepackage{amsmath}

\usepackage{wrapfig}
\usepackage{url}

\usepackage{hyperref}
\usepackage[dvips]{epsfig}
\usepackage{amsfonts}
\usepackage{amssymb,amsmath}
\usepackage{subfig}
\usepackage{cases}
\usepackage{multirow}
\usepackage{tikz}
\usepackage{epstopdf}
\usepackage{algorithmic}
\usepackage{psfrag}
\usepackage{cancel}
\usepackage{arydshln}
\usepackage{wasysym}
\usepackage{latexsym}
\usepackage{subfig}
\usepackage{float}
\usepackage[ruled,vlined]{algorithm2e}
\usepackage{lipsum}                     
\usepackage{xargs}
\usepackage{xcolor}
\usepackage{ulem}

\hypersetup{
    colorlinks,
    linkcolor={green},
    citecolor={red},
    urlcolor={blue}
}
\usepackage{mathtools,extarrows}
\usepackage[colorinlistoftodos]{todonotes}

\usepackage{pdflscape}
\usepackage{afterpage}
\usepackage{soul}
\usepackage{cite}
\usepackage[font=scriptsize]{caption}

\newcommand{\norm}[1]{\left\lVert#1\right\rVert}

\title{\Large \bf
Pareto Frontier Approximation Network (PA-Net) to Solve Bi-objective TSP
\vspace{-2 mm}
}

\author{Ishaan Mehta, Sharareh Taghipour, and Sajad Saeedi
\vspace{- 8 mm}
\thanks{\hspace*{-1.5em} Department of Mechanical and Industrial Engineering, Toronto Metropolitan University. Emails: {\tt \scriptsize \{ishaan.mehta, sharareh, s.saeedi\}@ryerson.ca} 
}
}

\begin{document}

\maketitle

\begin{abstract}
    \textcolor{black}{\textcolor{black}{The} travelling salesperson problem (TSP) is a classic resource allocation problem used to find an optimal order of doing a set of tasks while minimizing (or maximizing) an associated objective function. It is widely used in robotics for applications such as planning and scheduling. In this work, we solve TSP for two objectives using reinforcement learning (RL). Often in multi-objective  optimization problems, the associated objective functions can be conflicting in nature. In such cases, the optimality is defined in terms of Pareto optimality. A set of these Pareto optimal solutions in the objective space form a Pareto front (or frontier). Each solution has its trade-off. } 
   We present the Pareto frontier approximation network (PA-Net), a network that generates good approximations of the Pareto front for the bi-objective travelling salesperson problem (BTSP). Firstly, BTSP is converted into a constrained optimization problem. We then train our network to solve this constrained problem using the Lagrangian relaxation and policy gradient. \textcolor{black}{With PA-Net we improve the performance over an existing deep RL-based method. The average improvement in the hypervolume metric, which is used to measure the optimality of the Pareto front, is $2.3 \%$. At the same time, PA-Net has $4.5\times$ faster inference time.} Finally, we present the application of PA-Net to find optimal visiting order in a robotic navigation task/coverage planning. \textcolor{black}{Our code is available on the project website}\footnote{Project website: \href{https://sites.google.com/view/pa-net-btsp}{https://sites.google.com/view/pa-net-btsp}}.

\end{abstract}

\vspace{- 2 mm}
\section{Introduction} \label{sec:rw}

\par
\textcolor{black}{The} travelling salesperson problem (TSP) is a popular sequencing problem. TSP generates a sequence (a.k.a. tour) that visits each city (or node) in a given graph and finally returns to the starting city. The goal of TSP is to minimize the overall traversal cost of the tour. TSP and its variants are widely used in robotics for applications like path planning for UAVs \cite{xu2019brief}, multi-robot path planning \cite{yu2002implementation}, task allocation for robotic manipulators \cite{zacharia2005optimal}, and coverage planning \cite{bormann2018indoor}.

\par
We \textcolor{black}{use the} bi-objective travelling salesperson problem (BTSP) for application of coverage planning. Algorithms for coverage planning generate trajectories for robots to cover a given area \cite{bormann2018indoor}. Area coverage is used in robotic applications like  cleaning robots and surveillance. Grid-based TSP planners \cite{bormann2018indoor} segment a given map into multiple cells and generate a coverage pattern for each cell. The optimal visiting order for these cells is generated by solving TSP that minimizes the length of the tour. We are interested in a scenario where a robot has to visit these cells and the order is dependent on two objectives. The first objective is tour length, and the second objective could be used to represent traversable condition or priority of the path. So, BTSP would be an appropriate choice in this scenario. 
\par
There are a wide variety of algorithms, ranging from exact methods to evolution-based methods that solve multi-objective  TSP \cite{lust2010multiobjective}. Evolutionary algorithms, like non-dominated sorting genetic algorithm-II (NSGA-II) \cite{mtsp6} and multi-objective evolutionary algorithm (MOEA/D) \cite{mtsp7}, are frequently used to tackle multi-objective  TSP and other multi-objective  optimization problems. Some works use evolutionary algorithms coupled with local search heuristics  \cite{mtsp8,mtsp9,mtsp10}. In practice, these evolutionary-based methods suffer in performance and computation time with an increase in the scale of the problem \cite{zhang2016decision}. Another method is to use linear scalarization of the objectives using convex weights \cite{boyd2004convex}. These weights are used to convert BTSP to single objective TSP, which can be solved using solvers like \href{https://developers.google.com/optimization}{OR-Tools}. The downside of the linear scalarization technique is that it is unable to find solutions in the concave region of the Pareto front \cite{boyd2004convex}.


\par
\textbf{Contribution}: In this work, we address the aforementioned challenges. We present the Pareto frontier approximation network (PA-Net), a reinforcement learning (RL) based framework that generates an approximation of a set of Pareto optimal tours for BTSP. \textcolor{black}{Further, we demonstrate the use of PA-Net for a robotic planning application. Our main contributions are:}
 (1) \textcolor{black}{We achieve an average improvement of $2.3 \%$ in optimality metrics along with $4.5\times$ faster inference times as compared to the network (called DRL-MOA) proposed by Li \textit{et al.} \cite{li2020deep};}
 (2) \textcolor{black}{We address the drawback of the existing approach \cite{li2020deep}, which trains separate networks to generate different solutions for approximating the Pareto front. The generalization ability of PA-Net enables it to produce a dense approximation of the Pareto front through a single network, while maintaining similar training times as DRL-MOA \cite{li2020deep}}; and
 (3) Our approach can be extended to generate a set of Pareto optimal solutions for other multi-objective reinforcement learning and multi-objective optimization (MOO) tasks.

\par
\textbf{Related Work}: Sequencing problems are a subset of combinatorial optimization (CO) where the decision variables are discrete. Most CO problems are NP-Hard, and as a result state-of-the-art algorithms rely on handcrafted heuristics to make decisions that are otherwise too expensive to compute or  are tailored to specific problem instances. Recently, researchers are addressing these issues using deep learning and machine learning (ML) \cite{bengio2020machine,vesselinova2020learning,mazyavkina2020reinforcement}.
\par
\textcolor{black}{Many recent works of deep learning-based CO methods focus on solving Euclidean TSP.}  Google Brain's Pointer Network (Ptr-Net) \cite{deepTsp4} learns the conditional probability of an output sequence of elements that are discrete tokens corresponding to \textcolor{black}{the} positions in an input sequence. They used Ptr-Net to solve Euclidean TSP (and other CO problems) in an end-to-end fashion, where the solutions from classical methods are used as baselines for training. Similarly, TSP was solved using a 2D graph convolution network followed by the beam search procedure \cite{deepTsp1}. Deep reinforcement learning (DRL) was used to solve various combinatorial optimization problems in \cite{deepTsp3}. Their network uses RNN-based encoder and a Ptr-Net. They trained their network using policy gradient. Deudon \textit{et al.} \cite{deepTsp5} developed an actor-critic based architecture. They used transformers to encode the TSP graph. Further, they used 2-opt heuristics to refine the solutions generated by the network. Kool \textit{et al.} \cite{kool2018attention} proposed a network based on attention layers. They trained their network using policy gradient. Their network outperformed other deep-learning based solvers. Further, methods like \cite{deepTsp5,kool2018attention, ma2021learning} have been able to close the gap  (in terms of optimality) in comparison to TSP solvers like \href{https://www.math.uwaterloo.ca/tsp/concorde.html}{Concorde} and  \href{https://developers.google.com/optimization}{OR-Tools} by using additional heuristics or sampling procedures \cite{deepTsp5,kool2018attention}.

\par
Finding Pareto optimal solutions has  been studied in deep learning literature for various multi-objective tasks. In supervised learning  domain, many works focussed on multi-objective classification tasks \cite{pareto1,pareto2,pareto3,pareto4, navon2021learning}. Similarly, many multi-objective RL methods have been developed to solve multi-objective MDPs \cite{roijers2013survey,parisi2014policy}.  Some works in RL train many single policy networks to approximate the Pareto front \cite{MORL4,li2020deep}. While others have trained a single network to generate a set of Pareto optimal solutions \cite{ParetoRL,pi2}. 
\par
There are various methodologies to tackle MOO problems. The $\epsilon$-constrained methods optimize one of the objectives at a time, while using the other objectives as constraints \cite{mavrotas2009effective, chinchuluun2007survey}.  One of the most common ways is to use preference vectors or weights. These preference vectors or weights indicate the desired trade-off between various objectives.
Some methods use convex weights to scalarize the objective function, and the Pareto front can be obtained by solving the optimization for multiple preference vectors \cite{coello2009advances,boyd2004convex}. \textcolor{black}{On the other hand, some methods use these preferences as constraints. For instance, Das \textit{et al.} \cite{das1998normal} generate preferences (used in constraints) based on the convex hull of individual minima of the MOO. Our method also uses preference in constraints, although in our case preferences are unit vectors sampled from the unit circle in the objective space.} 

\par
Li \textit{et al.} \cite{li2020deep} solved multi-objective  TSP by training multiple single policy networks. They converted the MOO problem into a single objective using the linear scalarization method with convex preferences. They train multiple networks with the architecture adopted from \cite{nazari2018reinforcement}. Each network is trained with a different preference weight, to approximate the Pareto front. Their network, called DRL-MOA, generated competitive results in comparison to classical methods. However, the downside of their method is that it is redundant to train multiple networks, \textcolor{black}{which is time-consuming and resource-intensive}. Furthermore, solutions on concave regions of \textcolor{black}{the} Pareto front cannot be uncovered by their network because it uses the linear scalarization technique \cite{boyd2004convex}. In our work, we train a single network that can predict solutions for any unit preference vector. This enables us to produce a much denser Pareto front. Instead of using linear scalarization, our network learns to solve a constrained optimization problem where the constraints are dependent on the preference vectors.    


\par

\par
\textcolor{black}{In this work, we present PA-Net, a network framework trained using the policy gradient that can approximate the Pareto front for MOO problems. Our choice of using RL is motivated by the success of deep learning-based CO methods and the fact that it is hard to generate training data for complex problems like BTSP.  We use PA-Net to find a set of Pareto optimal tours for BTSP. The networks that use the policy gradient methods can be easily adopted and modified under our framework. So, the networks presented in \cite{deepTsp5,kool2018attention} are augmented for PA-Net to solve BTSP. The novelty of our algorithm is that we pose the problem of finding a set of Pareto optimal solutions as a constrained optimization problem, rather than using linear scalarization of the objective function. We use preference vectors as constraints which indicate the desired location of the solution in the objective space. Finally, we train our network using the reward constrained policy optimization, which is a paradigm used to train constrained RL applications \cite{tessler2018reward}. Our network performs better than other deep learning methods in terms of quality of the Pareto front, training, and inference time. Further, we extend our framework to process non-Euclidean data. We demonstrate the use of this network for a coverage planning application.}

\begin{figure}[t]
\vspace{- 2 mm}
    \centering
 \includegraphics[width = 0.8 \linewidth]{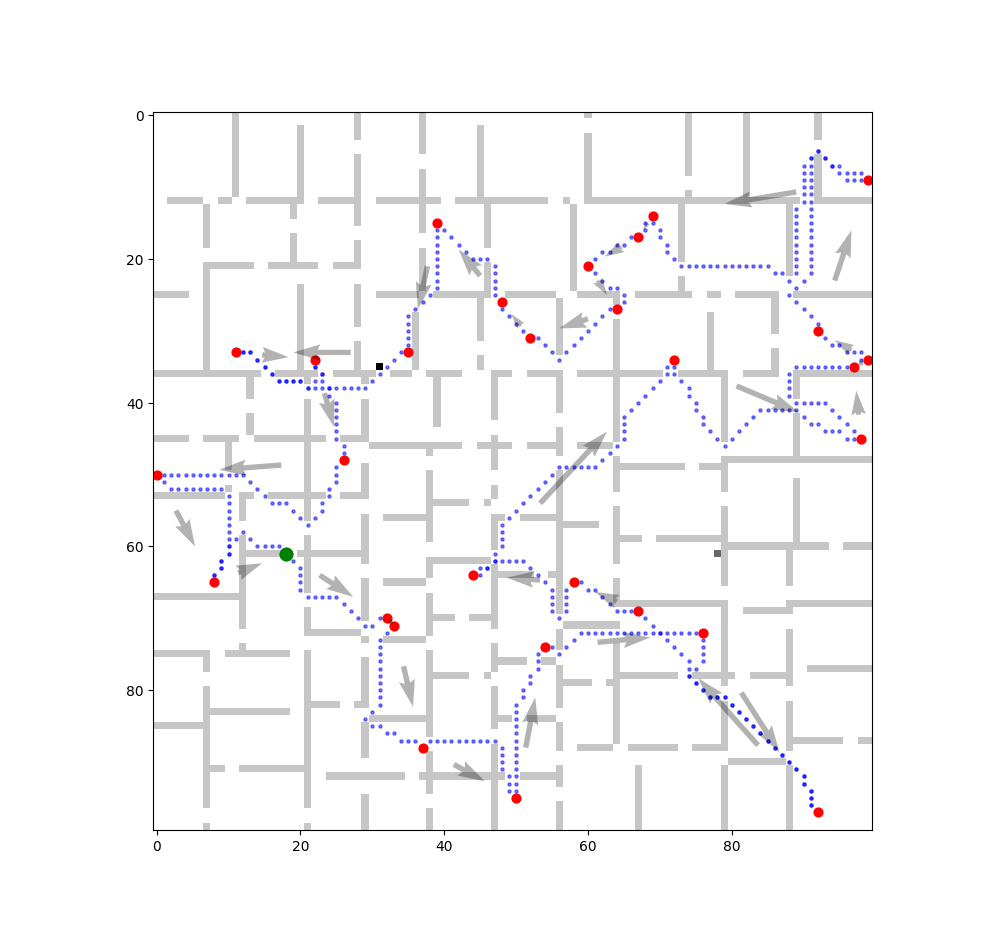}
    \caption{  A small scale BTSP tour generated by our algorithm for a map of a real-world environment. The locations of interest highlighted in red and the starting point in green. The path traversed by the robot is highlighted in blue, and arrows indicate the sequence of visiting different locations. It is clear that the path between two points of interest is not always a straight line. For such cases, we use a modified architecture of PA-Net, which can process adjacency matrices as inputs.
    \label{fig:viz0}} 
    \vspace{- 6 mm}
\end{figure}
\vspace{-2 mm}
\section{Background}\label{sec:background}
\par
Here we review the definition of Pareto optimality and present a brief primer on solving TSP using DRL. 
\vspace{-2.5 mm}
\subsection{Problem Setup}
\par
A MOO problem is defined as:
\small
\begin{equation}
min_{x} \; \vec{F}(x) = [f_{1}(x)\, f_{2}(x)\, ... \, f_{m}(x)]^{\top}\\
\end{equation}
\normalsize

where $\vec{F}(x)$ is a vector of m-objective functions and $x \in \textbf{X}$ is the vector of the decision variable in $\mathbb{R}^{n}$. In such problems, often different objectives are conflicting in nature, \textit{i.e.}, no single solution can simultaneously optimize all the objectives. Instead, a set of Pareto optimal solutions provide the best solutions with different trade-offs between various objectives. Pareto optimality is defined as follows:
\begin{itemize}
    \item \textbf{Dominance:} A solution $x^a$ is said to dominate $x^b$ ($x^a \prec x^b$) if and only if $f_{i}(x^a) \leq f_{i}(x^b)$, $\forall i \in \{1,...m\}$ and $f_{j}(x^a) < f_{j}(x^b)$ such that $\exists j \in \{1,...m\}$ .
     \item \textbf{Pareto optimality}: A solution $x^{*}$ is said to be Pareto optimal if there does not exist any solution $x'$ such that $x{'} \prec x^{*}$. A set of all such points form a Pareto frontier, denoted by \textbf{$\Upsilon$}.
\end{itemize}

\par
\textbf{Euclidean BTSP:} The Euclidean TSP is defined over a graph of $n$ cities, where each city has coordinates $a \in \mathbb{R}^2$. A TSP tour $\pi$ provides a sequence of visiting cities exactly once and then returning to the starting city. BTSP is a MOO problem that finds a set of Pareto optimal TSP tours $\Pi$ ($\Pi \subset \mathbb{Z}^n$) on a complete graph $s$, while optimizing for two objectives. Here, the input graph $s$ is a sequence of $n$ cities in a four-dimensional space $s = [a_i^1, a_i^2]_{i=1:n}$, where $a_i^m \in \mathbb{R}^2$ for each $m \in \{1,2\}$ \cite{li2020deep}. The goal is to find a tour $\pi \in \Pi$ that visits each city in the graph $s$ and can simultaneously optimize the objectives for $m \in \{1,2\}$: 
\vspace{-2.5 mm}
 \small
\begin{equation}\label{eq:tsp_obj}
     f_{m}(\pi) = \norm{a_{\pi(n)}^m -a_{\pi(1)}^m }_2 + \sum_{i=1}^{n-1} \norm{a_{\pi(i)}^m - a_{\pi(i+1)}^m }_2.
     \vspace{-2 mm}
\end{equation}
 \normalsize
\textbf{Non-Euclidean BTSP:} \textcolor{black}{In certain situations, the objective functions may not take a Euclidean form. For instance, the distance between two locations on a map might not to be a straight line because of obstacles or other path constraints, as shown in Fig.~\ref{fig:viz0}}. So for this case, we assume that the costs between all these points are given by the adjacency matrix, \textit{i.e.}, $\mathrm{\mathbf{H_m}}$ where $m \in \{1,2\}$. The cost associated with the $m^{th}$ objective for a tour $\pi$ can be calculated using:
\vspace{- 2.5 mm}
\small
\begin{equation}\label{eq:tsp_obj2}
     f_{m}(\pi) = \mathrm{\mathbf{H_m}}[\pi(n),\pi(1)] + \sum_{i=1}^{n-1} \mathrm{\mathbf{H_m}}[\pi(i),\pi(i+1)].
     \vspace{-2 mm}
\end{equation}
\normalsize

\subsection{TSP using policy gradient}
\par
\textcolor{black}{In our work, we adopt and modify architectures presented in \cite{kool2018attention}and \cite{deepTsp5}. These networks are trained using REINFORCE, a classic policy gradient method  \cite{williams1992simple}}. The input to the network is a graph $s$. An encoded representation of each city in the graph is obtained using an encoder. This encoded representation, along with the history of previous actions, are used by the decoder to sequentially generate a TSP tour. The actor, network $\theta$,  is trained to minimize the total tour length given by Eq.~\eqref{eq:tsp_obj} or Eq.~\eqref{eq:tsp_obj2}. The network is trained on a batch of  TSP problem instances of size $B$.  The training objective for the actor is given by:
\vspace{- 2 mm}
\small
\begin{equation}\label{eq:ac_obj}
    D(\theta) = \mathbb{E}_{s \sim \mathcal{S}}[\mathbb{E}_{\pi \sim p_{\theta}(.|s)}[Q(\pi | s)]].
    \vspace{- 2 mm}
\end{equation}
\normalsize
Here, $\mathcal{S}$ is the distribution from which training graphs are drawn and $p_{\theta}(\pi|s)$ is the probability of a tour generated by the decoder and $Q(\pi|s)$ is the reward which \textcolor{black}{minimizes} the total path length.

\vspace{-3 mm}
\section{Methodology}\label{sec:algo}
\par
This section provides our mathematical formulation for MOO. We use this formulation to train a network that generates a good approximation of the Pareto front for BTSP.
\vspace{-4 mm}
\subsection{Problem Formulation}
\vspace{-1 mm}
\par
\textcolor{black}{We intend to generate a good quality approximation of \textcolor{black}{the Pareto front, which is} denoted by $\widetilde{\Upsilon}$, where $\widetilde{\Upsilon} \subset \Upsilon$. Ideally, the set $\widetilde{\Upsilon}$ should capture a wide range of possible dominant solutions in the objective space.} 
\par
BTSP is an extension of TSP to the MOO domain. Let the vector cost function for a BTSP be given by $\vec{F}(\pi)$,  where $\vec{F} \in \mathbb{R}^2$ and the tour $\pi \in \Pi$ ($\Pi \subset \mathbb{Z}^n$). The optimization problem can be written as:

\small
\begin{equation}\label{eq:prob1}
   min_{\pi} \; \vec{F}(\pi) = [f_1(\pi) \, f_2(\pi)]^\top .
\end{equation} 
\normalsize

\textcolor{black}{Where the $i^{th}$ cost function is $f_i : \mathbb{Z}^n \rightarrow \mathbb{R}$} for $ i \in \{ 1,2\}$. We further assume that all the cost functions are strictly positive:
\small
\begin{equation}\label{eq:cstr1}
   f_{i}(\pi) > 0 \;\;\;\; \forall i \in \{ 1,2\}.
   \vspace{-1mm}
\end{equation}
\normalsize
\par
In order to find the Pareto front, we convert the MOO in Eq.~\eqref{eq:prob1} to a set of constrained optimization problems. \textcolor{black}{This is done by discretizing the objective space using a collection of $K$ unit preference vectors $W: \, \{\vec{w}_1...\vec{w}_K\}$, where each $ \vec{w}_k \, \in \, \mathbb{R}^2$ for $\forall k \in\{1...K\}$ . These preference vectors are a set of rays emanating from the origin that uniformly divide the objective space. Each element in $\vec{w}_k$ lies in the interval \textcolor{black}{$[0,1]$} and $||\vec{w}_k||_2 = 1$}.
\par
The key idea is to solve a surrogate optimization problem along each preference vector in order to generate a set of dominant solutions for Eq.~\eqref{eq:prob1}. This surrogate optimization is expressed as a set of $K$ constrained optimization problems, where the $k^{th}$ problem corresponding to $\vec{w}_k \in W$ is given by:
\small
\begin{equation} \label{eq:prob2}
\begin{aligned}
    min_{\vec{F}(\pi_k)} \; J(\vec{F}(\pi_k)) = ||{\vec{F}}(\pi_k)||_{2} \\
    s.t. \;\; \vec{F}(\pi_k) \in C_k . \\
\end{aligned}
\vspace{- 2 mm}
\end{equation}
\normalsize

\noindent The constraint set is defined as $C_k =\{g({\vec{F}(\pi)},\vec{w}_k) \leq 0 \; | \; \vec{F}(\pi) \in \mathbb{R}^2 \}$ and the corresponding tour set is defined as $A_k = \{g({\vec{F}(\pi)}, \vec{w_k}) \leq 0 \; | \; \pi \in \mathbb{Z}^n \}$. Here, the dot product constraint $g(\vec{F}(\pi_k), \vec{w}_k)$ is given by:
\vspace{-1.5 mm}
\small
\begin{equation}\label{eq:cstr2}
    g({\vec{F}(\pi_k)}, \vec{w}_k) = 1 - \frac{{\vec{w}}_k\cdot{\vec{F}}(\pi_k)}{||{\vec{F}}(\pi_k)||_{2}} .
\end{equation}
\vspace{-0.5 mm}
\normalsize
We assume $A_k$ is non-empty. As a consequence of this assumption, $C_k$ is also non-empty.

\begin{figure}[t]

    \centering
 \includegraphics[width = 0.7 \linewidth]{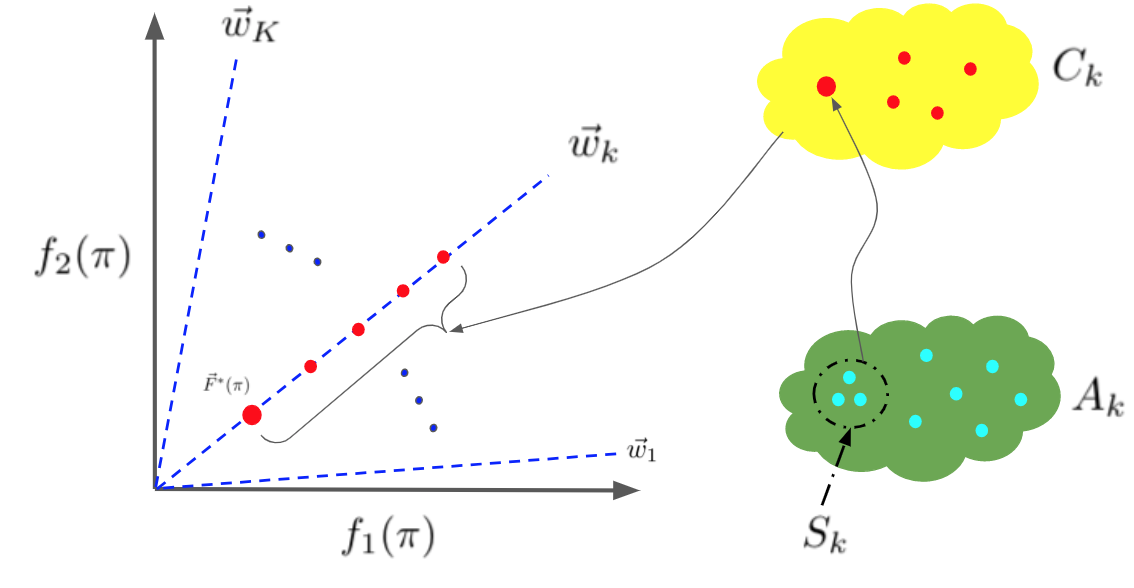}
    \caption{Visualization of surrogate optimization of Eq.~\eqref{eq:prob2} for 2-D cost ($\vec{F}(\pi)$) along the preference vector $\vec{w}_k$. All the members in the set $C_k$ are mapped to the line along the preference vector $\vec{w}_k$ in the objective space. The optimum point $\vec{F}^*(\pi)$ dominates all other members in $C_k$. The dominant point $\vec{F}^{*}$ is unique in $C_k$, although alternative solutions \textit{\textit{i.e.}} $\pi_k \in S_k$ may exist that map to $\vec{F}^{*}$. \label{fig:viz1}}
    \vspace{-6 mm}
\end{figure}

\textcolor{black}{The constraint set $C_k$ is a set of vector cost $\vec{F}(\pi_{k})$ where $\pi_k \in A_k$. Here, each $\vec{F}(\pi_{k})$  lies on the corresponding unit preference vector $\vec{w}_k$ in objective space. The objective function in Eq.~\eqref{eq:prob2} minimizes the $\mathbb{L}_2$-norm and hence finds the points closer to origin. Below, we state a theorem that is the motivating factor of our work.}

\par
\noindent \textbf{Theorem 1.} \textit{$\vec{F}^{*} \in C_k$ is the optimum solution of Eq.~\eqref{eq:prob2} if and only if it dominates all other points in the set.}

\par
\noindent \textbf{Proof:} Let $\vec{F}^{''} \in C_k$ minimize the Eq.~\eqref{eq:prob2} such that $\vec{F}^{*} \prec \vec{F}^{''}$. This dominance relation implies that \textcolor{black}{$f_{i}^{*} \leq f_{i}^{''} \; \forall i \in \{1,2\}$ and $\exists j \in \{1,2\}$} such that $f_{j}^{*} < f_{j}^{''} $ . This dominance relation leads to the following result:
\vspace{-1.5 mm}
\small
\begin{equation}
    ||\vec{F}^{*}||_2 < ||\vec{F}^{''}||_2 \, .
\end{equation}
\normalsize
\vspace{-0.5 mm}
\noindent But this result is a contradiction because $\vec{F}^{''}$ minimizes Eq.~\eqref{eq:prob2}. Hence, $\vec{F}^{*} \in C_k$ that dominates all other points in the set are the optimum solution for problem Eq.~\eqref{eq:prob2}. $\blacksquare$
\par
A similar argument can be made to show that a non-dominated solution in $C_k$ is the optimum solution of Eq.~\eqref{eq:prob2}. The dominant point in the set, \textit{i.e.}, $\vec{F}^{*} \in C_k$ is also unique in the set. An intuitive proof for this can be visualized using the case for $C_k \subset \mathbb{R}^2$ as shown in Fig.~\ref{fig:viz1}. Because of the dot product constraint, all the possible members in the set lie on the unit vector $\vec{w}_k$. It is clear from Fig.~\ref{fig:viz1} that the point in the set $C_k$ closest to origin dominates all other points and is, in fact, the optimum solution of Eq.~\eqref{eq:prob2}. Further, there can be multiple solutions in $A_k$ that lead to the dominant objective value, \textit{i.e.}, $\vec{F}^{*}$. Mathematically, the solution set  \textcolor{black}{$S_k \subset A_k$, where $S_k = \{\vec{F}(\pi_k) = \vec{F}^* \; | \; \pi_k \in A_k \}$}, all members of $S_k$ will generate dominant objective values.

\par
Essentially, Theorem 1 demonstrates the viability of approximating the Pareto front for the problem in Eq.~\eqref{eq:prob1} through the surrogate optimization problem in Eq.~\eqref{eq:prob2}. Solving the optimization problem in Eq.~\eqref{eq:prob2} for large values of $K$ can be computationally intractable.  We address this issue through the generalization power of deep neural networks. PA-Net learns to approximately solve Eq.~\eqref{eq:prob2} on a given input preference set. We select a sparse preference set (more details in Sec.~\ref{sec:NA}) that captures the diversity of the solutions in the objective space. This enables PA-Net to generalize to unseen preferences as well. Unlike  linear scalarization methods, our formulation can also find concave Pareto frontier. We demonstrate this with an example of a concave Pareto front in the Appendix. 
\vspace{-1 mm}
\subsection{Network Architecture}\label{sec:NA}
\par
\textbf{Euclidean BTSP:}
Our training framework can easily be used by any existing policy gradient-based networks to solve TSP. In this work, we adopt and modify the architecture from  existing works to solve Euclidean BTSP. The first network we use is Encode Attend Navigate (EAN) network \cite{deepTsp5}. The second network we use is the Attention network (AT) from \cite{kool2018attention}. We refer to these modified networks trained with our framework as \textbf{PA-EAN} and \textbf{PA-AT}, respectively. 

\par
\textcolor{black}{The input for Euclidean BTSP is $4D$ coordinates of the cities, where each pair of coordinates is used to calculate the corresponding objective.} To generate a set of dominant tours, we augment these network architecture by adding an input of a set of preference vectors $W$ of size $K$. Each $\vec{w}_k \in W$ is encoded in higher dimensions using a single layer feed-forward network. These additional layers learn features corresponding to different preferences. This encoding is combined with the encoded representation of the graph and then passed on to the decoder. With this architecture, the network can be trained for various preferences. 
\par
\textbf{Non-Euclidean BTSP:} \textcolor{black}{For the case where the input is an adjacency matrix, we are required to learn a representation of each city corresponding to each objective. This representation is used as an input to the network to generate a set of TSP tours.} A graph transformer-based encoder by Sykora \textit{et al.} \cite{sykora2020multi} is used to learn this representation. The input to the encoder is  the adjacency matrix $\mathrm{\mathbf{H_j}}$ and an initial set of features of each city in the graph, $s$ given by ${s}^j = [{a}^{j}_{1}...{a}^{j}_{n}]$ where $j \in \{1,2\}$. The description of the initial feature for the $i^{th}$ city corresponding to the $j^{th}$ objective, $a^{j}_{i} \in \mathbb{R}^3$, is given in Table~\ref{tab:fts}.
\vspace{- 3 mm}
\begin{table}[h]
    \centering
    \caption{Description of initial city features for PA-AD}
    \begin{tabular}{c c c}
    \hline
         Name & Dim. & Type  \\
         \hline
         Sum of neighbouring edge weights & 1  & float \\
         Min. neighbouring edge weight & 1  & float \\
         Max. neighbouring edge weight & 1  & float \\
         \hline
         
    \end{tabular}
    
    \label{tab:fts}
\vspace{-2.5 mm}
\end{table}
\par
The graph encoder produces an encoded representation of the features for each objective, \textit{i.e.}, $\Tilde{s}^j$. It should be noted that we process features for each objective independent of the other. Finally, we stack the learned features for each objective to obtain a combined representation  $\bold{\Tilde{S}} = [\Tilde{s}^1 \, \Tilde{s}^2]^{\top}$. Now, $\bold{\Tilde{S}}$ along with the preference encoding, is used by the network to predict BTSP tours. For this case, we use the AT network \cite{kool2018attention} along with the preference encoder. We will refer to this combined network as \textbf{PA-AD}. 

\par
\vspace{-2 mm}
\subsection{Training Methodology}

\par
The reward-constrained policy optimization is an actor-critic algorithm \textcolor{black}{that solves constrained RL problems} \cite{tessler2018reward}. It uses a Lagrangian of the constrained problem as  the objective function, where after each gradient update step, the Lagrangian multipliers are updated based on the constraint violation. We use the  reward-constrained policy optimization to train a network to solve the problem in Eq.~\eqref{eq:prob2} for all $k \in \{1...K\}$. 
\par
We intend to train a single network that generates a set of dominant tours $T: \{ \pi_1 ,..., \pi_K\}$. Hence, the problem in Eq.~\eqref{eq:prob2} for each $\pi_k \in T$ can be written in the parametric format as:
\vspace{-2mm}
\small
\begin{equation} \label{eq:prob3}
\begin{aligned}
    min_{\theta} \; J(\pi_{k}(\theta,s_j)) = ||{\vec{F}}(\pi_{k}(\theta,s_j))||_{2} \\
    s.t. \;\; g_{k}({\vec{F}(\pi_{k}(\theta,s_j),\vec{w}_k})) \leq 0.
\end{aligned}
\vspace{-1mm}
\end{equation}
\normalsize
\par
Here, $\theta$ is the parameters of the actor network and $\pi_k (\theta, s_j)$ is the tour generated by the actor network corresponding to $k^{th}$ preference for the input graph $s_j$. For notational convenience, we denote $\pi_k(\theta, s_j)$ as $\pi^{j}_{k}$. The Lagrangian dual problem for Eq.~\eqref{eq:prob3} is:
\vspace{-1mm}
\small
\begin{equation} \label{eq:prob4}
\begin{aligned}
  L_{k}(\pi^{j}_{k} , \lambda_k) \, = \, max_{\lambda_k \geq 0} \;  min_{\theta} \; J(\pi^{j}_{k} ) +  \lambda_k \cdot g_{k}({\vec{F}(\pi^{j}_{k}, \vec{w}_k})).
  \end{aligned}
  \vspace{-1mm}
\end{equation}
\normalsize
\par
Here, $\lambda_k$ is the $k^{th}$ Lagrangian multiplier corresponding  to the preference vector $\vec{w}_k$. We use the Lagrangian in Eq.~\eqref{eq:prob4} as the reward for the network. Based on this reward, the training objective for the actor  can be written in our case as:
\vspace{-1.5 mm}
\small
\begin{equation}
    D_{AC}(\theta) = \mathbb{E}_{s_j \sim S}[\mathbb{E}_{\vec{w}_k \sim W}[\mathbb{E}_{\pi \sim p_{\theta}(.|s_j)}[ L_{k}(\pi^{j}_{k} ,\lambda_k)]]].
\end{equation}
\normalsize
\vspace{-0.2 mm}
A critic network is used to provide predictions $b_{\phi}(\vec{w}_k,s_j)$ on the reward given in Eq.~\eqref{eq:prob4}. This critic network is trained on the mean squared error between its predictions and rewards of the actor, which is given by:
\vspace{-2.5 mm}
\small
\begin{equation}
    D_{CR}(\phi) = \frac{1}{K \cdot B} \sum_{k=1}^{K} \sum_{j=1}^{B}||b_{\phi}(\vec{w}_k,s_j)-  L_{k}(\pi^{j}_{k}  , \lambda_k) ||^{2}_{2}.
    \vspace{-2 mm}
\end{equation}
\normalsize

\par
\noindent The gradient for the training of the actor network is approximated using REINFORCE \cite{williams1992simple}:
\vspace{-1.1 mm}
\small
  \begin{equation} \label{eq:gradpa}
  \begin{aligned}
    \nabla_{\theta}D_{AC}(\theta) \approx \frac{1}{K \cdot B} \sum_{k=1}^{K} \sum_{j=1}^{B} [ L_{k}(\pi^{j}_{k}, \lambda_k) - \\ b_{\phi}(\vec{w}_k,s_j)]\cdot \nabla_{\theta}log(p_{\theta}(\pi^{j}_{k} ))].
   \end{aligned}
 \end{equation}
\normalsize

The description of the training of PA-Net is given in \textbf{Algorithm 1}. We start with the initialization of weights and learning rates for the network and the set of preference vectors, along with other hyperparameters that are the ascent rate of the Lagrangian multipliers $\alpha$ and $[\lambda_{min}, \lambda_{max}]$ the limits for the multipliers. The network is trained for $N$ iterations. At each iteration, a batch of graphs $\Omega$ of size $B$ is generated. For each $s_j \in \Omega$ and the corresponding preference vector, $\vec{w}_k \in W$ a tour $\pi^{j}_{k}$ is constructed by the network. Based on generated tours, the objective for actor and critic are calculated. Then parameters of the network are updated using gradient descent. At the end of each iteration, the Lagrangian multiplier corresponding to each preference vector is updated in an ascent step using:
\vspace{-2.5 mm}
\small
\begin{equation} \label{eq:lang_update}
     \lambda^{i+1}_k = \Gamma_{\lambda}(\lambda^{i}_k + \frac{\alpha}{B} \cdot \sum_{j=1}^{B} g_{k}({F(\pi^{j}_{k},\vec{w}_k}))).
\end{equation}
\normalsize
\vspace{-4 mm}
\par
Here $\Gamma_{\lambda}(\cdot)$ ensures that the multipliers remain within the limits, \textit{i.e.}, $[\lambda_{min}, \lambda_{max}]$ and $\alpha$ is prespecified ascent rate. Although the network is trained on a fixed set of preferences $W$, it can generalize to a larger set of preferences. \textcolor{black}{For training of BTSP, we generate preferences by sampling the unit circle in $\mathbb{R}^2$. Since the objective functions are strictly positive, the angle for unit vectors is sampled from the interval $(0^{\circ}, 90^{\circ})$. A higher number of preferences, \textit{i.e.}, would lead to better performance. But increasing $K$ also increases training times. So, to keep the training tractable, we limit the number to $K=20$. } 
\par
One downside of our approach is that our network might be susceptible to local minima. However, in practice, our approach can generate competitive results as presented next.
\vspace{- 6 mm}
\begin{algorithm}[h]
\scriptsize
\SetKwData{Left}{left}\SetKwData{This}{this}\SetKwData{Up}{up}
\SetKwFunction{Union}{Union}\SetKwFunction{FindCompress}{FindCompress}
\SetKwInOut{Input}{input}\SetKwInOut{Output}{output}
\BlankLine
\Input{ $[\theta , \phi, \eta_A , \eta_C], [W, \alpha ,\lambda_{1:K}, \lambda_{min}, \lambda_{max}] \, \xleftarrow{} \,$ Initialization of network weights, set of preference vectors and corresponding parameters. }
\Output{Trained network parameters of PA-Net $\theta^* , \phi^*$. }
\BlankLine
 \For{$i \, \xleftarrow{} 1,2...N$}{
  $\Omega : \{s_1 \, .... \,s_B\} \leftarrow $ Sample a Batch of TSP Graphs of size $B$ from distribution $\mathcal{S}$.
  \BlankLine
  \For{$k \, \xleftarrow{} 1,2...K$}   
  {
    \For{$j \, \xleftarrow{} 1,2...B$} 
    {
    $\pi^{j}_k \, \xleftarrow{} \,$  Actor network Generates TSP Tour for each $s_j$ and $\vec{w}_k$.
    \BlankLine
    {$b_{\phi}(\vec{w}_k,s_j) \, \xleftarrow{} \,$ Critic Network predicts the baseline}
    \BlankLine
    $L_{k}(\pi^{j}_k,\lambda_k) \, \xleftarrow{} \,$ Calculate the Lagrangian using  Eq.~\eqref{eq:prob4}.
    }
  }
  \textbf{Actor Update}: $ \theta \, \xleftarrow{} \, \theta - \eta_A \cdot \nabla_{\theta}D_{AC}(\theta)$
  \BlankLine
  {\textbf{Critic Update}: $ \phi \, \xleftarrow{} \, \phi - \eta_C \cdot \nabla_{\phi}D_{cr}(\phi) $}

  \BlankLine
  \For{$k \, \xleftarrow{} 1,2...K$}
  {
  $ \lambda_k \, \xleftarrow{} \, $ Update the Lagrangian multipliers using Eq.~\eqref{eq:lang_update}
  }

 }
 \caption{Training of PA-Net}
 \normalsize
\end{algorithm}

\begin{figure*}[h]
\vspace{-6 mm}
\hspace{0.05 \linewidth}
\subfloat[40 City BTSP ]{\includegraphics[width=0.38\textwidth]{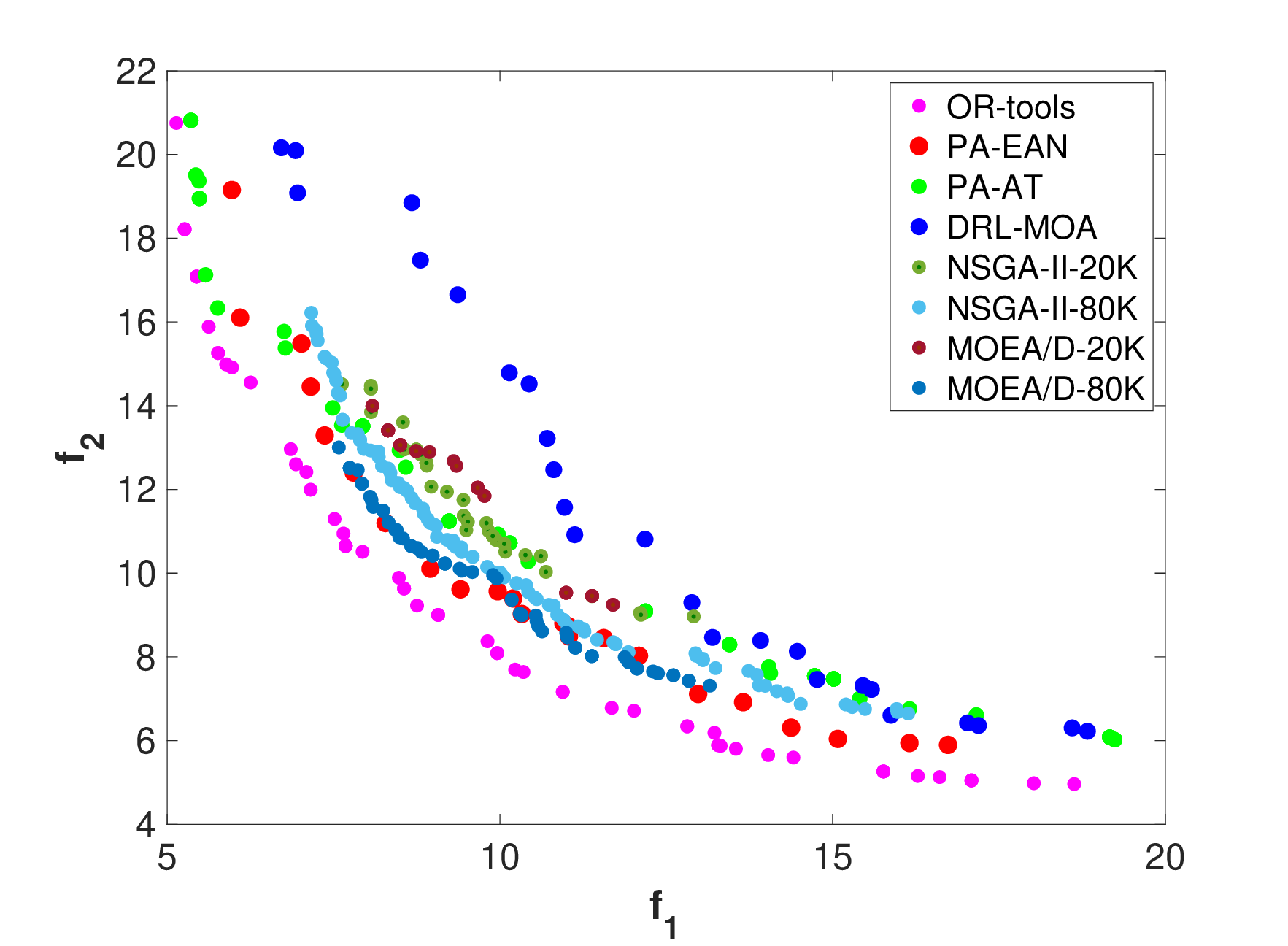}} 
\hspace{0.1\linewidth}
\subfloat[500 City BTSP]{\includegraphics[width=0.38\textwidth]{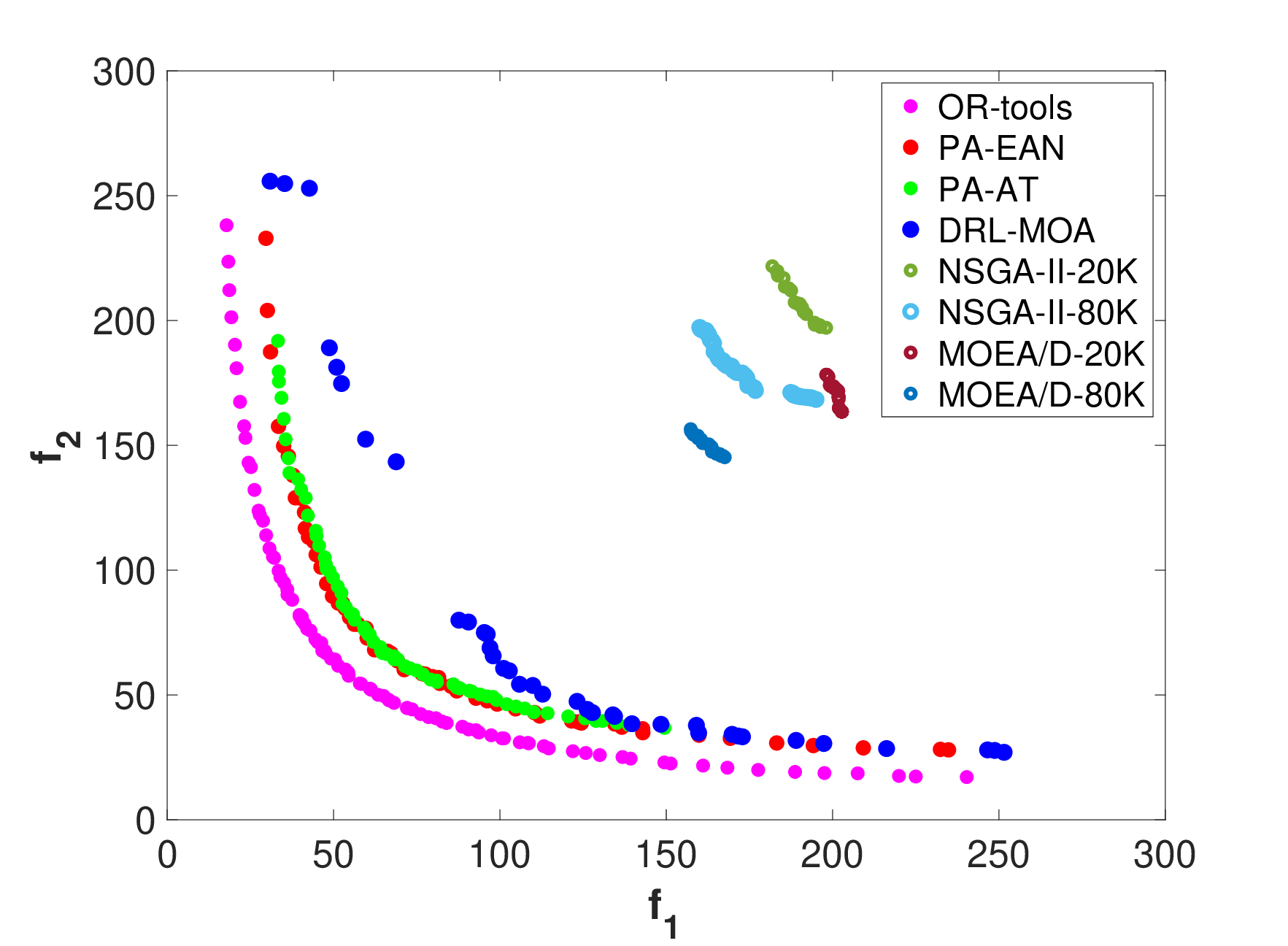}} \\
\vspace{0.01 \linewidth}
\hspace{0.3 \linewidth}
\vspace{-3mm}
\caption{Visualization of the dominant solutions for different problem instances. It can be seen that our network (PA-EAN and PA-AT) generates significantly better objective values than DRL-MOA and evolutionary methods \label{fig:Plots}}
\end{figure*}
\vspace{-8 mm}
\section{Experiments} \label{sec:expm}


\par
To evaluate the efficacy of PA-Net, we present two experiments. The first experiment is on BTSP instances, where the input is Euclidean data. The second experiment is an application for coverage planning, where we test our network on non-Euclidean input data generated from simulated grid world maps. 
\par
\textcolor{black}{The performance of PA-Net is compared with deep learning-based method DRL-MOA \cite{li2020deep}, evolution-based strategies: NSGA-II and MOEA/D \cite{mtsp7}. Further, we generate baseline results using TSP solver from \href{https://developers.google.com/optimization}{OR-Tools} library. Note that in this case, we use linear scalarization of objectives. The experiments for deep learning methods are carried out on NVIDIA V100 Volta GPU. Whereas, for NSGA-II, MOEA/D and OR-Tools experiments are carried out on dual-core Intel i5 processor.}

\par
The results of PA-Net are compared with other methods on the basis of hypervolume (HV) of the solutions. HV is a hybrid metric used to evaluate Pareto fronts \cite{mtsp3,audet2020performance}. It represents the volume covered by the non-dominated set of solutions with respect to the worst-case solution. A higher HV indicates a better quality of the Pareto front, both in terms of optimality and coverage of the objective space. Here, HV is calculated as the percentage of points dominated by the solution set out of densely sampled points in a fixed volume in the objective space. \textcolor{black}{ This volume is generated with respect to a reference point. We use a fixed reference point to compute HV for all the algorithms. It is calculated as the product of the number of cities and unity vector, for example, reference point for 200-city BTSP is $[200.0, 200.0]^{\top}$.}
\par
We also mention the run time of each algorithm. However, evolutionary-based methods and OR-Tools are CPU-based methods, whereas DRL-MOA and PA-Net are GPU-based methods. Libraries like OR-tools are highly optimized and can generate solutions in a short time. On the other hand, deep-learning based methods have relatively higher run-time, but they can take advantage of parallelization and generate a batch of solutions in one shot.

\begin{table}[h]
\vspace{-2 mm}
 \scriptsize
\caption{Training details for our networks and DRL-MOA. Our network can learn to generalize for various preferences for similar training time to DRL-MOA \vspace{-3mm} }
\label{tab:Results5}
\begin{center}
\begin{tabular}{c c c c c}
\hline 
 & PA-EAN & PA-AT & PA-AD & DRL-MOA\\
\hline 
Batch Size  &   $60$ &  $60$ & $60$ & $200$  \\
Epochs  &   $1$ &  $3$ & $4$ & $1/$net. \\
Steps (per epoch)& $20000$ & $5000$ &  $5000$ & $2500$ \\
Input Graph Size &   $ 120 \times 4$ &   $ 40 \times 4$ & $ 40 \times 40$  & $ 40 \times 4$   \\
Training Times (hrs) &   $\sim 12$ &  $\sim 8$ & $\sim 10$ & $\sim 9$  \\
\hline
\end{tabular}
\end{center}
\normalsize
\vspace{-4 mm}
\end{table}

\par

\begin{table*}[h]
 \scriptsize
 \caption {Quantitative comparison of solutions for BTSP. Our methods outperform DRL-MOA and other evolutionary algorithms in terms of HV and time.  } \label{tab:Results1}
 \vspace{-2 mm}
\begin{center}
\begin{tabular}{ c c c c c c c c c }
\hline
    & \multicolumn{2}{c}{\textbf{40-City}}
            & \multicolumn{2}{c}{\textbf{200-City}}
                    & \multicolumn{2}{c}{\textbf{500-City}}
                            & \multicolumn{2}{c}{\textbf{1000-City}}            \\
    
\hline
 Algo.  &   HV ($\%$)  &  Time (s)  &   HV ($\%$)  &   Time (s)   &   HV ($\%$)  &  Time (s)   &   HV ($\%$) &   Time (s)    \\ 
\hline
NSGA-II (20K)   &   67.3  &   5.64  &   45.4  &   8.04  &   38.4 &   14.7 &   33.8  &   27.1  \\

NSGA-II (80K)   &   72.5   &   21.7    &  53.9   & 30.8  &    46.36  &   58.7 &  41.48 & 107.3     \\

MOEA/D (20K)   & 66.7   & 9.357   &  47.5   &    12.7   &    40.4   &  20.7     &   35.7    &  33.5     \\

MOEA/D (80K)   & 70.7   &   34.65  &   55.98 &    48.2   &    48.8   &   79.53    &   43.89    &  130.75    \\
\hdashline
DRL-MOA   &     73.6  &   5.87    &    80.63   &   29.3    &    84.5   &  72.9     &   85.9 &    145.5   \\
 
PA-EAN (ours) &   { 75.4 } &    { \bf 1.59}   &    {83.2}   &  { \bf 6.03 }  & {  86.92 }   &  { 1\bf 5.14 }   &  {  88.45}   & { \bf 30.38 }   \\

PA-AT (ours) &   { 74.18 } &    {2.6}   &    { 80.9}   &  {  13.4 }  & { 85.15 }   &  { 33.5 }   &  {  87.35}   & {  68.2 }   \\
\hdashline
OR-tools ($n_l$ = NO LIMIT)   &     \bf 78.14  &   2.16    &    \bf 86.5   &   86.8    &    \bf 91.07   &  732    &   \bf 93.39 &    3730   \\
OR-tools ($n_l$ = 10)   &     78.08  &   0.93    &     84.8   &   10.4   &    89.8  &  53.5   &    92.38 &    209   \\
\hline
\end{tabular}
\end{center}
\normalsize
\vspace{-2 mm}
\end{table*}
\begin{table*}[h]
 \scriptsize
\caption {Quantitative comparison of Pareto front for Coverage Planning.Our algorithm generates promising initial results.} \vspace{-2 mm}
\label{tab:Results3}
\begin{center}
\begin{tabular}{ c c c c c c c}
\hline
    & \multicolumn{2}{c}{\textbf{40-City}}
            & \multicolumn{2}{c}{\textbf{100-City}}
                    & \multicolumn{2}{c}{\textbf{200-City}}
                                  \\
    
\hline
 Algo.  &   HV ($\%$)  &  Time (s)  &   HV ($\%$)  &   Time (s)   &   HV ($\%$)  &  Time (s)      \\ 

\hline
PA-AD (ours) &   { 71.59} &    {3.06}   &    { 72.14}   &  {  6.26 }  & { 76.4 }   &  { 11.85 }   \\

\hdashline
OR-tools ($n_l$ = 100)   &     75.9 &   2.15   &    78.7  &   17.4   &    84.29  &  82.5   \\
OR-tools ($n_l$ = 10)   &     66.6  &   0.75    &     68.5   &   2.72  &    82.4  &  9.9   \\
\hline
\end{tabular}
\vspace{-6 mm}
\end{center}
\normalsize
\end{table*}
\textbf{Training Setup:} For Euclidean BTSP, we train two architectures, \textit{i.e.}, PA-AT and PA-EAN. For the non-Euclidean BTSP, we train PA-AD. Training details of these networks along with DRL-MOA are given in Table~\ref{tab:Results3}. The training times are reported based on training on NVIDIA V100 Volta GPU. 
All the networks except PA-EAN are trained on 40-city BTSP instances. Further, all our networks are trained for $K = 20$ preferences. PA-AD network is trained on adjacency matrices.

\vspace{-2 mm} 
\subsection{Evaluation on Euclidean BTSP Instances}\label{sec:expma}
\par
In this experiment, we evaluate the performance of PA-Net on bi-objective TSP. The input graph $G$ is a sequence of $n$ cities, where each city $a_i \in \mathbb{R}^{4}$ is represented by $4$ dimensional features. We use $\mathbb{L}_2$ norm using Eq.~\eqref{eq:tsp_obj} for the two objectives. For each algorithm, we generate 100 solutions corresponding to various preferences among the two objectives. It should be noted that the preferences used in linear scalarization are sampled from [0,1] and satisfy convexity constraints. In PA-Net, the preferences used are unit vectors that uniformly segment the objective space. \par
A set of $25$ synthetically generated BTSP instances are used in this experiment. The average HV and run times for each algorithm are reported in Table~\ref{tab:Results1}. NSGA-II and MOEA/D are tested at different values for the maximum number of iterations. For OR-Tools, we report results for different values of the maximum number of iterations for solution refinement, $n_l$, where the solver terminates once the specified limit is reached. The results of the quantitative comparison for BTSP is given in Table~\ref{tab:Results1}.  Visualization of dominant objective values attained by different algorithms is shown in Fig.~\ref{fig:Plots} (a)-(b).
\par
\textbf{Discussion:} It is clear that the baseline method \textit{i.e.}, OR-Tools achieves the best values for HV metric. Further, these values improve with increasing the maximum solution limit of the solver. The routing problem solver of OR-Tools is a highly optimized library built on years of research by the operation research community. \textcolor{black}{It should be mentioned, that even the state-of-the-art deep learning-based methods to solve single objective TSP are able to compete against OR-Tools by using additional neural local heuristics or sampling a large set of solution space \cite{kool2018attention,deepTsp5,ma2021learning}.} On the other hand, evolutionary methods significantly underperform as the scale of the problem increases. This is likely because these algorithms are unable to explore the solution space well for larger problems. Running these algorithms for more iterations could potentially improve their performance in terms of HV, but this comes with an additional computational cost. All deep learning-based methods achieve competitive results in terms of HV. Our networks achieve much better performance as compared to DRL-MOA in terms of HV. \textcolor{black}{The average improvement in HV over DRL-MOA for PA-EAN and PA-AT is $2.3 \%$ and, $0.7 \%$ respectively. Further, PA-EAN and PA-AT generates the complete Pareto front about $4.5 \times$ and $2.1 \times$ faster, respectively, as compared to DRL-MOA. Further, we have either lower or comparable training times relative to DRL-MOA, see Table~\ref{tab:Results5}.} For each problem set, PA-Net can infer a solution from a single network, whereas DRL-MOA has to train and rely on multiple networks. Another notable point is that DRL-MOA has many gaps in its Pareto front, see Fig.~\ref{fig:Plots}. This demonstrates the ineffectiveness of linear scalarization approach in deep learning setting. 
\vspace{-2 mm}
\subsection{Application for Coverage Planning}

\par
\textcolor{black}{We use BTSP for the application of coverage planning. In this experiment, the robot has to visit various locations in a given map while optimizing for total distance travelled and an additional priority metric. An example of  such a map is shown in Fig.~\ref{fig:viz0}.  Such a priority metric can be representative of attributes like traversable conditions of the path or traffic on a given path \textit{etc}. The distances between locations are stored in adjacency matrices for this case because the distances are not Euclidean.}
\par
For this experiment, we synthetically generate $10$ maps using PathBench \cite{toma2021pathbench}, an open-source motion-planning benchmarking platform. We then randomly sample points from the free space in the map as locations of interest, as shown in Fig.~\ref{fig:viz0}. We assume that the graph formed by these points are fully connected. The input adjacency matrix for the first objective, $\mathrm{\mathbf{H_1}}$, is generated by determining path-lengths between all the points using A* algorithm. For the second objective, we randomly sample $2D$ points where each coordinate is in range [0,1). We then generate the second adjacency matrix $\mathrm{\mathbf{H_2}}$, by computing the Euclidean distance between these points. For this dataset, we test the performance of our PA-AD network while using OR-Tools as the baseline. The average results for HV and runtimes are reported in Table~\ref{tab:Results3}.
\par
\textbf{Discussion:} OR-Tools finds much better solutions when this limit is increased. As mentioned before, OR-Tools is a highly optimized solver for TSP and other routing problems. Like the Euclidean case, there is a lag in the performance of our network as compared to OR-Tools. \textcolor{black}{This is likely because the network converges to local minima. Nonetheless, these initial results look encouraging as we are able to outperform existing deep learning approach, \textit{i.e.}, DRLMOA. }

\vspace{-2 mm}
\section{Conclusions}\label{sec:conclusion}
\par
We presented PA-Net, a network that approximates the Pareto frontier for the bi-objective TSP. Our results indicate a competitive performance in terms of optimality of the solutions. This is achieved by segmenting the objective space using a set of unit vectors which represent trade-offs among various objectives. We then use these preference vectors to convert the unconstrained optimization problem into a set of constrained optimization problems. Then the network is trained using Lagrangian relaxation and policy gradient to generate solutions for these constrained problems. While PA-Net is trained on a fixed number of preference vectors, it generalizes well to other unseen preferences as well. The effectiveness of our method is highlighted by the significant gains made in terms of quality of solutions, inference and training times. Although we focus on bi-objective TSP in this work, our training framework can be applied to other MOO problems. We also demonstrated a use case of PA-Net for a coverage planning application. \textcolor{black}{Our future investigation would be focussed on bridging the performance gap with OR-Tools through informed use of local heuristics and tackling the issue of convergence to local minima by incorporating expert training data from solvers like OR-Tools and Concorde. Further, we also plan on extending this work to cases with dynamic costs and multi-robot systems.} 

\vspace{-2 mm} 
\section*{Appendix}
\vspace{-1 mm}
\subsection*{\textbf{Ablation Studies}}
\par
\vspace{-1 mm}
In this ablation study, we evaluate the contribution of the preference layer. We train two networks, \textit{\textit{i.e.}}, PA-EAN$^{-}$ and PA-AT$^{-}$. The training conditions for both these networks are kept the same as our proposed networks, with the only difference being that the preference embedding layer is removed for both these networks. We compare our proposed network with these on the basis of hypervolume. The results are reported in Table~\ref{tab:abl1}. It can be clearly seen that without the preference layer, the performance of the networks drop significantly.
\vspace{-1 mm}
\begin{table}[h]
    \centering
    \caption{Impact of using preference encoder. It can be seen that preference encoders play a critical role for the performance of our networks. }
    \scriptsize
    \begin{tabular}{c c c c c}
    
    \hline
         Network & 40-City & 200-City & 500-City & 1000-City  \\
         \hline
         & HV(\%) & HV(\%) & HV(\%) & HV(\%) \\
         \hline
         PA-EAN & { 75.4} & 83.2 & 86.92 & 88.4 \\
         PA-EAN$^-$ & 39.1& 25.6 & 25 & 24.6 \\
         \hdashline
         PA-AT & 74.18 & 80.9 & 85.15 & 87.3 \\
         PA-AT$^-$ & 52& 48.9 & 52.9 & 55.8 \\
         \hline
         
    \end{tabular}
    \normalsize
    \label{tab:abl1}
    \vspace{-2 mm}
\end{table}
\par
In the case of non-Euclidean BTSP, a key addition is the graph transformer layers that learn representation for initial features. In order to evaluate its impact, we train a network, \textit{i.e.}, PA-AD$^-$ without a graph convolution layer. In this case, we use the initial feature set described in Table~\ref{tab:fts} along with the coordinates of the city as the input to the network. We then compare PA-AD and PA-AD$^-$ on the map dataset generated for the non-Euclidean BTSP experiment. The results for comparison are reported in Table~\ref{tab:abl2}. It can be clearly seen that graph transformer encoder plays a crucial role to learn useful representations of features for each city.
\vspace{-6 mm}
\begin{table}[h]
    \centering
    
    \caption{Impact of using graph transformer encoder. Using a graph encoder enables the network to learn better features as indicated by higher HV values.}
    \scriptsize
    \begin{tabular}{c c c c}
    \hline
         Network & 40-City & 100-City & 200-City \\
         \hline
         PA-AD & 71.59 & 72.14 & 76.4  \\
         PA-AD$^-$ & 60 & 53 & 53.2 \\
         \hline
    \end{tabular}
    \normalsize
    \label{tab:abl2}

\end{table}
\vspace{-6 mm}
\subsection*{\textbf{Convergence to Concave Pareto Fronts}}
\begin{figure}[t]
    \centering
   
    \includegraphics[width = 0.55\linewidth]{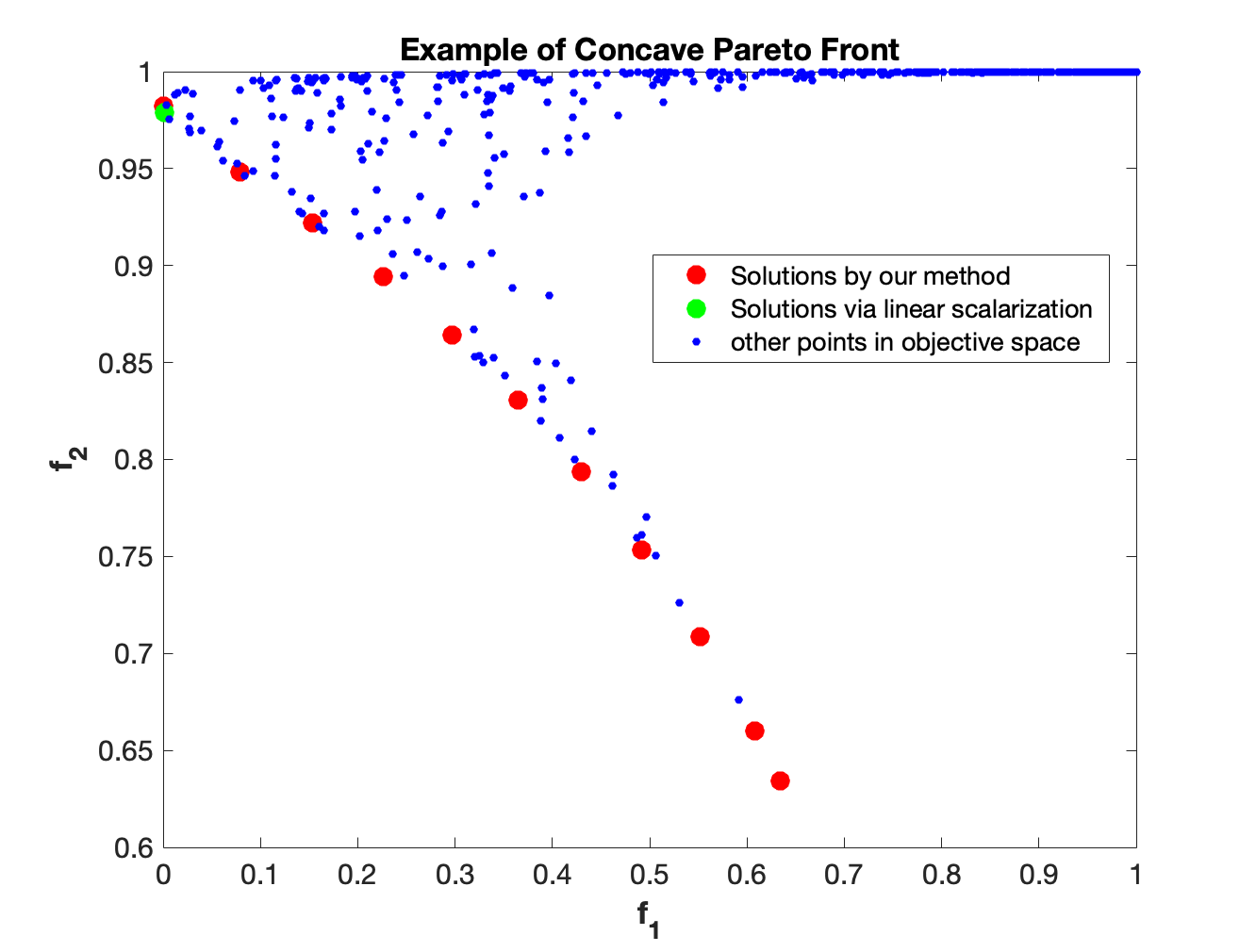}
    \caption{  \scriptsize{The above plot depicts the convergence of our method to Concave Pareto Front. Here red points are the results generated by our algorithm, green points are the solution from linear scalarization method and blue points represent the set of possible solutions in the objective space. }
    }\label{fig:my_label}
\vspace{- 8 mm}
\end{figure}
\par
We solve a concave MOO problem to demonstrate the convergence of our optimization framework given by Eq.~\eqref{eq:prob2}. The concave problem taken from \cite{pareto2} is given by:
 \small
\begin{equation}
min_x \, \vec{F}(x) = [f_1 (x), f_2(x)]^{\top},
\end{equation}
\normalsize
where,
\vspace{-2 mm}
 \small
\begin{equation}
\begin{aligned}
f_1(x) = 1 - exp(- \Sigma_{i = 1}^{d} (x_i - \frac{1}{\sqrt{d}})^2), \\
f_2(x) = 1 - exp(- \Sigma_{i = 1}^{d} (x_i + \frac{1}{\sqrt{d}})^2).
\end{aligned}
\end{equation}
\vspace{-2 mm}
\normalsize
\par
Here $x = [x_1, x_2]^{\top} \in \mathbb{R}^{2+}$ and $d=2$. The surrogate optimization in this case with preference $\vec{w_k}$  is given by:
\vspace{-2 mm}
 \small
\begin{equation}
\begin{aligned}
min_{\vec{F}(x^k)} \; J(\vec{F}(x^k)) = ||\vec{F}(x^k)||_2 \\
s.t. \;  1 - \frac{{\vec{w}}_k\cdot{\vec{F}}(x^k)}{J(\vec{F}(x^k))} \leq 0
\end{aligned}
\vspace{-2 mm}
\end{equation}
\normalsize
\par
We solve the above problem in Matlab
for $K=20$. The preference is generated by $\vec{w_k} = [cos(\phi_k), sin(\phi_k)]^{\top} $, where $\phi_k \in (0,90)$. We also solve the MOO problem with a simple linear scalarization of objective. In this case, the preference is given by $\alpha^k = [\alpha_1, \alpha_2]^{\top} \in \mathbb{R}^{2+}$ such that $\alpha_1 + \alpha_2 = 1$. We use $K=100$ preferences in this case. The $k^{th}$objective function for linear scalarization is:
\vspace{-2mm}
 \small
\begin{equation}
 min_{\vec{F}(x^k)} \; R(\vec{F}(x^k)) = \alpha_1 \cdot f_1(x^k) + \alpha_2 \cdot f_2(x^k)
 \vspace{-1mm}
\end{equation}
\normalsize
\par 
The results are shown in Fig.~\ref{fig:my_label}. It can be clearly seen that our method is able to produce the concave Pareto front. On the other hand, linear scalarization is unable to find solutions on the concave part of the Pareto front. This example demonstrates that our method can certainly be extended to MOO with concave Pareto fronts. 
\vspace{-1.5 mm}
\bibliographystyle{IEEEtran}
\bibliography{IEEEabrv,bib}

\begin{thebibliography}{10}
\providecommand{\url}[1]{#1}
\csname url@samestyle\endcsname
\providecommand{\newblock}{\relax}
\providecommand{\bibinfo}[2]{#2}
\providecommand{\BIBentrySTDinterwordspacing}{\spaceskip=0pt\relax}
\providecommand{\BIBentryALTinterwordstretchfactor}{4}
\providecommand{\BIBentryALTinterwordspacing}{\spaceskip=\fontdimen2\font plus
\BIBentryALTinterwordstretchfactor\fontdimen3\font minus
  \fontdimen4\font\relax}
\providecommand{\BIBforeignlanguage}[2]{{%
\expandafter\ifx\csname l@#1\endcsname\relax
\typeout{** WARNING: IEEEtran.bst: No hyphenation pattern has been}%
\typeout{** loaded for the language `#1'. Using the pattern for}%
\typeout{** the default language instead.}%
\else
\language=\csname l@#1\endcsname
\fi
#2}}
\providecommand{\BIBdecl}{\relax}
\BIBdecl

\bibitem{xu2019brief}
Y.~Xu and C.~Che, ``A brief review of the intelligent algorithm for traveling
  salesman problem in {UAV} route planning,'' in \emph{2019 IEEE 9th
  International Conference on Electronics Information and Emergency
  Communication (ICEIEC)}.\hskip 1em plus 0.5em minus 0.4em\relax IEEE, 2019,
  pp. 1--7.

\bibitem{yu2002implementation}
Z.~Yu, L.~Jinhai, G.~Guochang, Z.~Rubo, and Y.~Haiyan, ``An implementation of
  evolutionary computation for path planning of cooperative mobile robots,'' in
  \emph{Proceedings of the 4th World Congress on Intelligent Control and
  Automation (Cat. No. 02EX527)}, vol.~3.\hskip 1em plus 0.5em minus
  0.4em\relax IEEE, 2002, pp. 1798--1802.

\bibitem{zacharia2005optimal}
P.~T. Zacharia and N.~Aspragathos, ``Optimal robot task scheduling based on
  genetic algorithms,'' \emph{Robotics and Computer-Integrated Manufacturing},
  vol.~21, no.~1, pp. 67--79, 2005.

\bibitem{bormann2018indoor}
R.~Bormann, F.~Jordan, J.~Hampp, and M.~H{\"a}gele, ``Indoor coverage path
  planning: Survey, implementation, analysis,'' in \emph{2018 IEEE
  International Conference on Robotics and Automation (ICRA)}.\hskip 1em plus
  0.5em minus 0.4em\relax IEEE, 2018, pp. 1718--1725.

\bibitem{lust2010multiobjective}
T.~Lust and J.~Teghem, ``The multiobjective traveling salesman problem: a
  survey and a new approach,'' in \emph{Advances in Multi-Objective Nature
  Inspired Computing}.\hskip 1em plus 0.5em minus 0.4em\relax Springer, 2010,
  pp. 119--141.

\bibitem{mtsp6}
B.~A. Beirigo and A.~G. dos Santos, ``Application of {NSGA-II} framework to the
  travel planning problem using real-world travel data,'' in \emph{2016 IEEE
  Congress on Evolutionary Computation (CEC)}.\hskip 1em plus 0.5em minus
  0.4em\relax IEEE, 2016, pp. 746--753.

\bibitem{mtsp7}
W.~Peng, Q.~Zhang, and H.~Li, ``Comparison between moea/d and {NSGA-II} on the
  multi-objective travelling salesman problem,'' in \emph{Multi-objective
  memetic algorithms}.\hskip 1em plus 0.5em minus 0.4em\relax Springer, 2009,
  pp. 309--324.

\bibitem{mtsp8}
A.~Jaszkiewicz, ``On the performance of multiple-objective genetic local search
  on the 0/1 knapsack problem-a comparative experiment,'' \emph{IEEE
  Transactions on Evolutionary Computation}, vol.~6, no.~4, pp. 402--412, 2002.

\bibitem{mtsp9}
L.~Ke, Q.~Zhang, and R.~Battiti, ``Hybridization of decomposition and local
  search for multiobjective optimization,'' \emph{IEEE transactions on
  cybernetics}, vol.~44, no.~10, pp. 1808--1820, 2014.

\bibitem{mtsp10}
X.~Cai, Y.~Li, Z.~Fan, and Q.~Zhang, ``An external archive guided
  multiobjective evolutionary algorithm based on decomposition for
  combinatorial optimization,'' \emph{IEEE Transactions on Evolutionary
  Computation}, vol.~19, no.~4, pp. 508--523, 2014.

\bibitem{zhang2016decision}
X.~Zhang, Y.~Tian, R.~Cheng, and Y.~Jin, ``A decision variable clustering-based
  evolutionary algorithm for large-scale many-objective optimization,''
  \emph{IEEE Transactions on Evolutionary Computation}, vol.~22, no.~1, pp.
  97--112, 2016.

\bibitem{boyd2004convex}
S.~Boyd, S.~P. Boyd, and L.~Vandenberghe, \emph{Convex optimization}.\hskip 1em
  plus 0.5em minus 0.4em\relax Cambridge university press, 2004.

\bibitem{li2020deep}
K.~Li, T.~Zhang, and R.~Wang, ``Deep reinforcement learning for multiobjective
  optimization,'' \emph{IEEE Transactions on Cybernetics}, 2020.

\bibitem{bengio2020machine}
Y.~Bengio, A.~Lodi, and A.~Prouvost, ``Machine learning for combinatorial
  optimization: a methodological tour d’horizon,'' \emph{European Journal of
  Operational Research}, 2020.

\bibitem{vesselinova2020learning}
N.~Vesselinova, R.~Steinert, D.~F. Perez-Ramirez, and M.~Boman, ``Learning
  combinatorial optimization on graphs: A survey with applications to
  networking,'' \emph{IEEE Access}, vol.~8, pp. 120\,388--120\,416, 2020.

\bibitem{mazyavkina2020reinforcement}
N.~Mazyavkina, S.~Sviridov, S.~Ivanov, and E.~Burnaev, ``Reinforcement learning
  for combinatorial optimization: A survey,'' \emph{arXiv preprint
  arXiv:2003.03600}, 2020.

\bibitem{deepTsp4}
O.~Vinyals, M.~Fortunato, and N.~Jaitly, ``Pointer networks,'' in
  \emph{Advances in neural information processing systems}, 2015, pp.
  2692--2700.

\bibitem{deepTsp1}
C.~K. Joshi, T.~Laurent, and X.~Bresson, ``An efficient graph convolutional
  network technique for the travelling salesman problem,'' \emph{arXiv preprint
  arXiv:1906.01227}, 2019.

\bibitem{deepTsp3}
I.~Bello, H.~Pham, Q.~V. Le, M.~Norouzi, and S.~Bengio, ``Neural combinatorial
  optimization with reinforcement learning,'' \emph{arXiv preprint
  arXiv:1611.09940}, 2016.

\bibitem{deepTsp5}
M.~Deudon, P.~Cournut, A.~Lacoste, Y.~Adulyasak, and L.-M. Rousseau, ``Learning
  heuristics for the tsp by policy gradient,'' in \emph{International
  conference on the integration of constraint programming, artificial
  intelligence, and operations research}.\hskip 1em plus 0.5em minus
  0.4em\relax Springer, 2018, pp. 170--181.

\bibitem{kool2018attention}
W.~Kool, H.~Van~Hoof, and M.~Welling, ``Attention, learn to solve routing
  problems!'' \emph{arXiv preprint arXiv:1803.08475}, 2018.

\bibitem{ma2021learning}
Y.~Ma, J.~Li, Z.~Cao, W.~Song, L.~Zhang, Z.~Chen, and J.~Tang, ``Learning to
  iteratively solve routing problems with dual-aspect collaborative
  transformer,'' \emph{Advances in Neural Information Processing Systems},
  vol.~34, 2021.

\bibitem{pareto1}
O.~Sener and V.~Koltun, ``Multi-task learning as multi-objective
  optimization,'' in \emph{Advances in Neural Information Processing Systems},
  2018, pp. 527--538.

\bibitem{pareto2}
X.~Lin, H.-L. Zhen, Z.~Li, Q.-F. Zhang, and S.~Kwong, ``Pareto multi-task
  learning,'' in \emph{Advances in Neural Information Processing Systems},
  2019, pp. 12\,060--12\,070.

\bibitem{pareto3}
D.~Mahapatra and V.~Rajan, ``Multi-task learning with user preferences:
  Gradient descent with controlled ascent in pareto optimization,'' in
  \emph{International Conference on Machine Learning}.\hskip 1em plus 0.5em
  minus 0.4em\relax PMLR, 2020, pp. 6597--6607.

\bibitem{pareto4}
M.~Ruchte and J.~Grabocka, ``Efficient multi-objective optimization for deep
  learning,'' \emph{arXiv preprint arXiv:2103.13392}, 2021.

\bibitem{navon2021learning}
\BIBentryALTinterwordspacing
A.~Navon, A.~Shamsian, G.~Chechik, and E.~Fetaya, ``Learning the pareto front
  with hypernetworks,'' in \emph{International Conference on Learning
  Representations}, 2021. [Online]. Available:
  \url{https://openreview.net/forum?id=NjF772F4ZZR}
\BIBentrySTDinterwordspacing

\bibitem{roijers2013survey}
D.~M. Roijers, P.~Vamplew, S.~Whiteson, and R.~Dazeley, ``A survey of
  multi-objective sequential decision-making,'' \emph{Journal of Artificial
  Intelligence Research}, vol.~48, pp. 67--113, 2013.

\bibitem{parisi2014policy}
S.~Parisi, M.~Pirotta, N.~Smacchia, L.~Bascetta, and M.~Restelli, ``Policy
  gradient approaches for multi-objective sequential decision making,'' in
  \emph{2014 International Joint Conference on Neural Networks (IJCNN)}.\hskip
  1em plus 0.5em minus 0.4em\relax IEEE, 2014, pp. 2323--2330.

\bibitem{MORL4}
P.~Vamplew, R.~Issabekov, R.~Dazeley, C.~Foale, A.~Berry, T.~Moore, and
  D.~Creighton, ``Steering approaches to pareto-optimal multiobjective
  reinforcement learning,'' \emph{Neurocomputing}, vol. 263, pp. 26--38, 2017.

\bibitem{ParetoRL}
R.~Yang, X.~Sun, and K.~Narasimhan, ``A generalized algorithm for
  multi-objective reinforcement learning and policy adaptation,'' in
  \emph{Advances in Neural Information Processing Systems}, 2019, pp.
  14\,636--14\,647.

\bibitem{pi2}
S.~Parisi, M.~Pirotta, and M.~Restelli, ``Multi-objective reinforcement
  learning through continuous pareto manifold approximation,'' \emph{Journal of
  Artificial Intelligence Research}, vol.~57, pp. 187--227, 2016.

\bibitem{mavrotas2009effective}
G.~Mavrotas, ``Effective implementation of the $\varepsilon$-constraint method
  in multi-objective mathematical programming problems,'' \emph{Applied
  mathematics and computation}, vol. 213, no.~2, pp. 455--465, 2009.

\bibitem{chinchuluun2007survey}
A.~Chinchuluun and P.~M. Pardalos, ``A survey of recent developments in
  multiobjective optimization,'' \emph{Annals of Operations Research}, vol.
  154, no.~1, pp. 29--50, 2007.

\bibitem{coello2009advances}
C.~C. Coello, C.~Dhaenens, and L.~Jourdan, \emph{Advances in multi-objective
  nature inspired computing}.\hskip 1em plus 0.5em minus 0.4em\relax Springer,
  2009, vol. 272.

\bibitem{das1998normal}
I.~Das and J.~E. Dennis, ``Normal-boundary intersection: A new method for
  generating the pareto surface in nonlinear multicriteria optimization
  problems,'' \emph{SIAM journal on optimization}, vol.~8, no.~3, pp. 631--657,
  1998.

\bibitem{nazari2018reinforcement}
M.~Nazari, A.~Oroojlooy, L.~Snyder, and M.~Tak{\'a}c, ``Reinforcement learning
  for solving the vehicle routing problem,'' \emph{Advances in neural
  information processing systems}, vol.~31, 2018.

\bibitem{tessler2018reward}
C.~Tessler, D.~J. Mankowitz, and S.~Mannor, ``Reward constrained policy
  optimization,'' \emph{arXiv preprint arXiv:1805.11074}, 2018.

\bibitem{williams1992simple}
R.~J. Williams, ``Simple statistical gradient-following algorithms for
  connectionist reinforcement learning,'' \emph{Machine learning}, vol.~8, no.
  3-4, pp. 229--256, 1992.

\bibitem{sykora2020multi}
Q.~Sykora, M.~Ren, and R.~Urtasun, ``Multi-agent routing value iteration
  network,'' in \emph{International Conference on Machine Learning}.\hskip 1em
  plus 0.5em minus 0.4em\relax PMLR, 2020, pp. 9300--9310.

\bibitem{mtsp3}
J.-F. B{\'e}rub{\'e}, M.~Gendreau, and J.-Y. Potvin, ``An exact phi-constraint
  method for bi-objective combinatorial optimization problems: Application to
  the traveling salesman problem with profits,'' \emph{European journal of
  operational research}, vol. 194, no.~1, pp. 39--50, 2009.

\bibitem{audet2020performance}
C.~Audet, J.~Bigeon, D.~Cartier, S.~Le~Digabel, and L.~Salomon, ``Performance
  indicators in multiobjective optimization,'' \emph{European journal of
  operational research}, 2020.

\bibitem{toma2021pathbench}
A.-I. Toma, H.-Y. Hsueh, H.~A. Jaafar, R.~Murai, P.~H. Kelly, and S.~Saeedi,
  ``{P}ath{B}ench: A benchmarking platform for classical and learned path
  planning algorithms,'' in \emph{2021 18th Conference on Robots and Vision
  (CRV)}.\hskip 1em plus 0.5em minus 0.4em\relax IEEE, 2021, pp. 79--86.

\end{thebibliography}
\end{document}